\documentclass[10pt,twocolumn,letterpaper]{article}

\usepackage{cvpr}              % To produce the CAMERA-READY version
\usepackage[accsupp]{axessibility} % Improves PDF readability for those with disabilities.

\usepackage{amsmath}
\usepackage{amssymb}
\usepackage{booktabs}
\usepackage{pifont}
\usepackage{color}
\usepackage[table]{xcolor}
\usepackage{bm}
\usepackage{multirow}
\usepackage{soul}
\usepackage{bbm}
\usepackage{subcaption}
\usepackage{xurl}

\usepackage{graphicx}
\usepackage{subcaption}
\usepackage{rotating}
\usepackage{svg}

\newlength{\subfigwidth}

\definecolor{cvprblue}{rgb}{0.21,0.49,0.74}
\usepackage[pagebackref,breaklinks,colorlinks,citecolor=cvprblue]{hyperref}

\usepackage{dutchcal} % \mathcal: also small letters. \mathbcal = bold ones. 
\usepackage{bbding} % for cross mark

\def\ours{\textbf{PALADIN}}

\def\paperID{15} % *** Enter the Paper ID here
\def\confName{ICCV}
\def\confYear{2025}

\title{\ours{} : Robust Neural Fingerprinting for Text-to-Image Diffusion Models}

\author{Murthy L$^{1,2}$ \thanks{Work done during part-time masters at IISc.}
\quad{Subarna Tripathi$^{1}$}\\
$^{1}$Intel Corporation \quad\quad $^{2}$Indian Institute of Science\\
{\tt\small murthy.l@intel.com \quad\quad\quad subarna.tripathi@intel.com}
}

\begin{document}
\maketitle

\begin{abstract}
The risk of misusing text-to-image generative models for malicious uses, especially due to the open-source development of such models, has become a serious concern. 
As a risk mitigation strategy, 
attributing generative models with neural fingerprinting is emerging as a popular technique. There has been a plethora of recent work that aim for addressing neural fingerprinting. A trade-off between the attribution accuracy and generation quality of such models has been studied extensively. None of the existing methods yet achieved $100\%$ attribution accuracy. However,  
any model with less than cent percent accuracy is practically non-deployable. In this work, we propose an accurate method to incorporate neural fingerprinting for text-to-image diffusion models leveraging the concepts of cyclic error correcting codes from the literature of coding theory. 
% Source code will be available.  

\end{abstract} 

\section{Introduction}
\label{sec:intro}

\quad{} The current state-of-the-art text-to-image generative models have shown significant capabilities in generating high-quality and diverse images from natural language text prompts. This has enabled the widespread creation and manipulation of photo realistic images, adhering to precise textual prompts. This has led to many image editing tools \cite{couairon2022diffeditdiffusionbasedsemanticimage} \cite{gal2022imageworthwordpersonalizing} \cite{ruiz2023dreamboothfinetuningtexttoimage} that are becoming creative resources for artists, designers, and the public.
Although this is a great leap forward for generative modeling, it raises concerns about the authenticity, veracity, validity, and provenance of the generated images \cite{url:AI_Generated_Artwork}. The increasing realism of such generated images 
% generated by generative AI 
poses a serious threat to both individual and social interests. A convincing and widely shared misinformation could disrupt democratic discussion, influence elections, or threaten national security. A viable solution to this problem is to create accountability for the generated image by integrating user information in the generated artifacts. 

A line of work in this area has emerged that analyze design principles of fingerprinting strategies in text-to-image diffusion models and the trade-off between the attribution accuracy and generated image quality. However, none of such methods is able to achieve 100\% accuracy. For example, an attribution accuracy of 99.9\% means the aforesaid method will still predict $1\ million$  user incorrectly for a database containing $1\ billion$ users, such methods can not be deployed in the real world. 

\begin{figure}[htbp]
\includegraphics[width=0.45\textwidth]{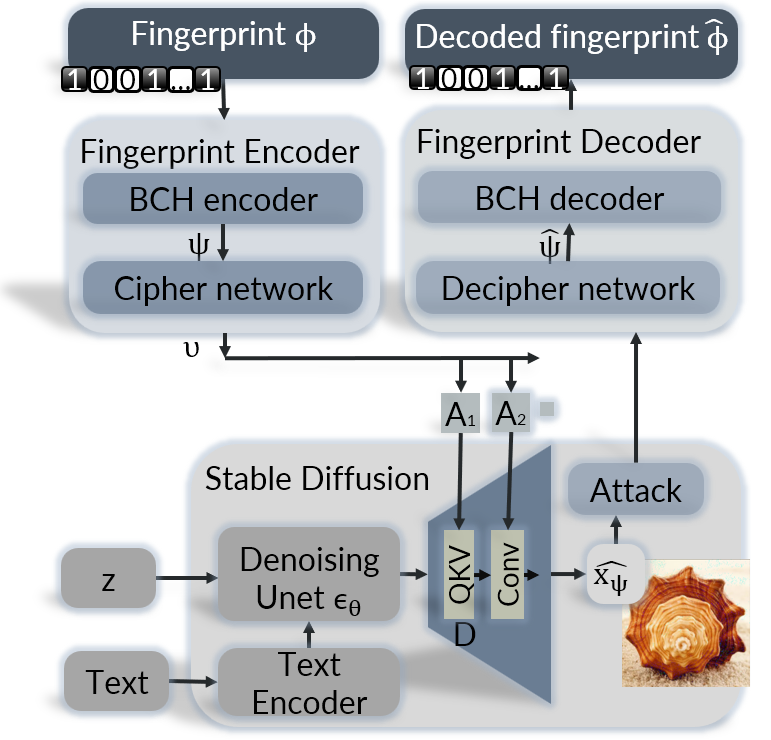}
\caption{\ours{} training pipeline. The weights of the denoising unet and text encoder are frozen and not part of training, weights of Cipher network, decoder part of Stable Diffusion and decipher network are updated as part of training. The attack block performs image post-processing such as Brightness, Gaussian Noise, H Flip, and others randomly with a probability of 50\%. BCH encoder and decoder block is configured to codeword length = 63, message length = 39, capable of detecting and correcting up to 4-bit errors.}
\label{fig:neural_fingerprinting_overview}
\end{figure}

% \subsection{Overview}

\setlength{\subfigwidth}{36mm}
\begin{figure*}[h]

\centering
\begin{tabular}{r p{\subfigwidth} p{\subfigwidth} r p{\subfigwidth} p{\subfigwidth}}

\begin{turn}{90} \quad\quad\quad\quad WOUAF\end{turn} & 
\begin{subfigure}[b]{\subfigwidth}
    \caption*{Generated Image (1)}
    \includegraphics[width=\subfigwidth]{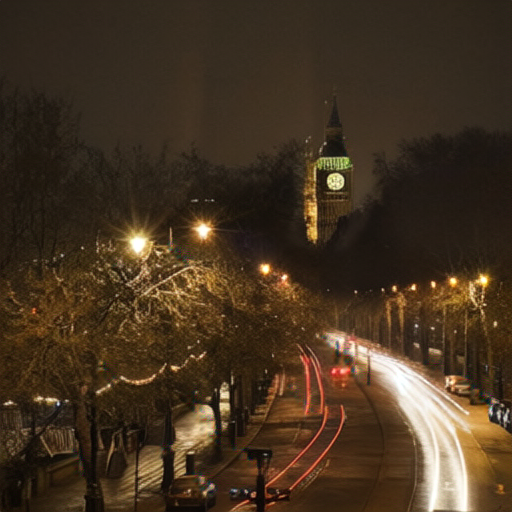}
\end{subfigure} &
\begin{subfigure}[b]{\subfigwidth}
    \caption*{Pixel wise Difference 2x (1) }
    \includegraphics[width=\subfigwidth]{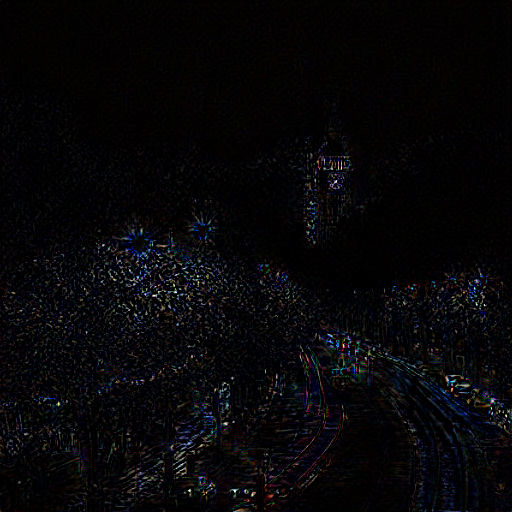}
\end{subfigure} &
 & 
\begin{subfigure}[b]{\subfigwidth}
    \caption*{Generated Image (2)}
    \includegraphics[width=\subfigwidth]{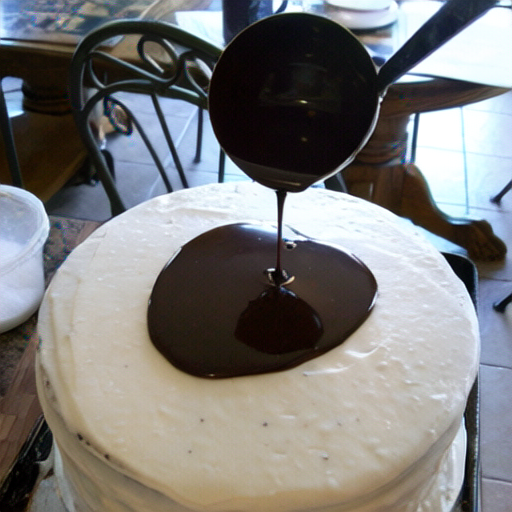}
\end{subfigure} &
\begin{subfigure}[b]{\subfigwidth}
    \caption*{Pixel wise Difference 2x (2)}
    \includegraphics[width=\subfigwidth]{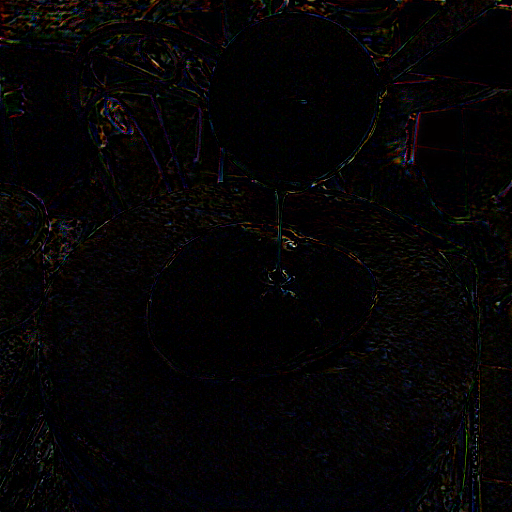}
\end{subfigure} \\

\begin{turn}{90} \quad\quad\quad\quad PALADIN\end{turn} & 
\includegraphics[width=\subfigwidth]{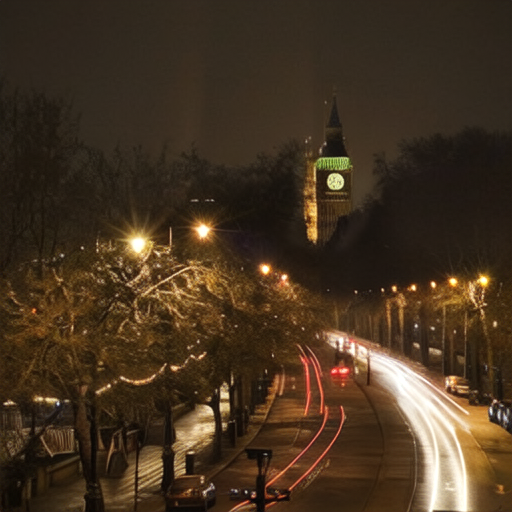} &
\includegraphics[width=\subfigwidth]{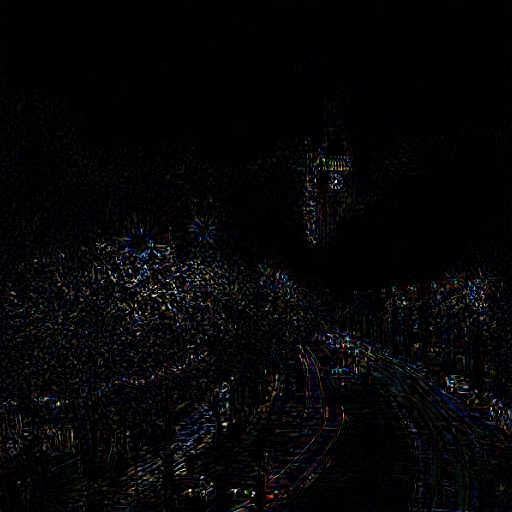} &
 & 
\includegraphics[width=\subfigwidth]{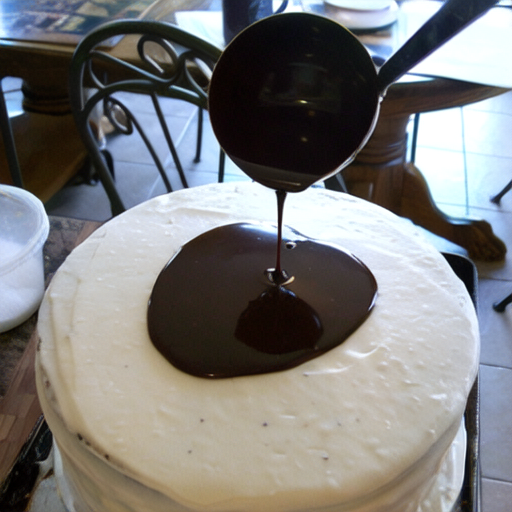}  &
\includegraphics[width=\subfigwidth]{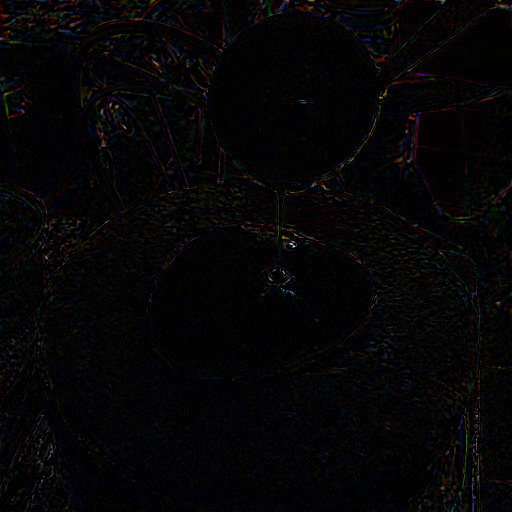} 
\end{tabular}
\caption{A qualitative comparison of PALADIN with WOUAF on COCO dataset. Pixelwise difference being measured w.r.t original image and amplified by 2x. PALADIN shows superior accuracy with no significant image residual. Prompts used 1) 'A city street at night with a tall tower at the end of a road.' and 2) 
'This cake is being doused in liquid chocolate'.
}
\label{fig:qualitative-results}
\end{figure*}

% \input{sec/imgs/qualitative}

% In this work, we explore and improvise neural fingerprinting for text-to-image diffusion models. 
Named after legendary nights and defender of a noble cause, we
introduce the \ours{} (\textbf{P}erfect user \textbf{A}ttribution for \textbf{LA}tent \textbf{DI}ffusio\textbf{N}) framework. \ours{} leverages cyclic error correcting codes to address the robust neural fingerprinting for latent diffusion models. It is capable of elevating a \emph{near-perfect} fingerprinting in latent diffusion models to be a \emph{perfect} one.
\ours{} is capable of handling image post-processing and detect responsible user for generated artifacts with $100\%$ confidence and also appropriately flagging non-identifiable fingerprints. 

Our implementation is based atop a state-of-the-art method, WOUAF~\cite{kim2024wouafweightmodulationuser}. However, the core principle can be applied to other models as well. 
To summarize, below is the list of our contributions.  
\begin{itemize}

\item 
First, we propose to improve both the image generation quality and attribution accuracy on top of our strongest baseline, WOUAF,  with a combination improvised module architecture and loss functions. 

\item
We propose a framework, \ours{}, for accurate fingerprinting 
% within text to image diffusion models 
% to elevate a \emph{near-perfect} model to a \emph{perfect} one
utilizing cyclic error correction codes. 

\item
Finally, we introduce FER (Fingerprint Error Rate), an informative and accurate metric to measure the effectiveness of such models. We show the efficacy of the model by extensive experiments on standard benchmarks.

\end{itemize}

% To the best of our knowledge, this is the first method that achieves 100\% attribution accuracy and the lowest FER for the text-to-image diffusion model. 

\section{Related Work}
\label{sec:related_work}
% We begin with the literature of model attribution with fingerprint encoding and decoding. 
\quad Yu et al. \cite{yu2022artificialfingerprintinggenerativemodels} proposed a weight modulation technique that modulates the generator's weights to embed user information in the generated artifacts. By incorporating fingerprints in the generator parameter rather than generator input facilitates in training a generic model that can be instantiated with unique fingerprint catering to a large user base. Despite these advancements, the above method caters to GAN based models and their suitability and effectiveness for diffusion based models remained unexplored. Zhang et al. \cite{zhang2024attackresilientimagewatermarkingusing} proposed a fingerprinting technique for latent diffusion models that embed fingerprinting in the Fourier space of the latent vector. However, this method is restricted to a single watermark per training cycle. Fernandez et al. \cite{fernandez2023stablesignaturerootingwatermarks} proposed a fingerprint embedding technique which embeds user information into generated artifacts by finetuning the latent decoder. A pretrained fingerprint encoder-decoder network is used to finetune the decoder to embed user information in the generated artifacts. However, this approach requires one train cycle per user, necessitating high compute and time. Kim et al. \cite{kim2024wouafweightmodulationuser} proposed a weight modulation technique in which weights of the Stable Diffusion (SD) are modulated to embed user information in the generated artifacts. This facilitates the generation and distribution of 
% multiple 
fingerprint decoders,
capable of handling multiple embeddings, 
% capable decoders 
with a single training cycle. Although this method achieves satisfactory results against attacks such as image augmentation
at the cost of robustness and probabilistic attribution accuracy 
% , but attribution accuracy is probabilistic and not robust enough, which hinders its use in any deployable 
rendering it un-deployable in
real-world scenarios for tracking the malicious user responsible for generating artifacts. 

One of the recent state-of-the-art method ,WOUAF~\cite{kim2024wouafweightmodulationuser}, 
introduced the concept of embedding the watermarks in all the images they
generate
by fine-tuning of the decoder of Latent Diffusion
Models. 
WOUAF~\cite{kim2024wouafweightmodulationuser} is based on StyleGAN \cite{karras2020analyzingimprovingimagequality} weight modulation technique to embed user information in the generated artifacts.
These watermarks are more robust than previously existing models could generate, invisible to the human eye and
could be employed to detect generated images and identify the
user that generated it, with high performance. 
% We make use of recent progress in fingerprinting in diffusion models in \ours{}. 
In ours{}, we make use of recent progress in fingerprinting in diffusion models and enhance their effectiveness with the power of error correction capability. 
\ours{} can broadly be classified as distributor-oriented fingerprinting, in which the model distributor assigns and embeds a unique traceable fingerprint to each user who downloads the model. 
In particular, we build atop WOUAF \cite{kim2024wouafweightmodulationuser}.  
However, the core principle can be applied to other models as well.

% However, our neural fingerprinting can be classified as inventor-oriented fingerprinting, in which the model inventor implements and embeds the unique information in the generated artifacts as part of training process, and distributor-oriented fingerprinting, in which the model distributor assigns and embeds a unique traceable fingerprint to each user who downloads the model. 

\section{Approach}
%Bit Err. Detect
\begin{table}
    \centering
        \setlength{\tabcolsep}{16pt}
        \resizebox{\linewidth}{!}{
        % \begin{tabular}{llll}
        \begin{tabular}{l|ccc}
        \toprule
         Methods    & Err. Detection & Bit Acc.($ \uparrow $) & FER ($ \downarrow $)\\
            \midrule
            WOUAF          & \XSolidBrush & 0.9974  & 0.0700*\\
            \ours{} $\psi$  & \XSolidBrush & 0.9999  & 0.0020*\\ 
            \ours{}  $\phi$ & \CheckmarkBold & \textbf{1.0000}  & \textbf {0.0004\ \ }  \\
        \bottomrule
    \end{tabular}
    }
    \caption{A comparison of fingerprint accuracy and fingerprint error rate (FER). Experiments were conducted on 512 x 512 images on MS-COCO validation dataset, with 32bit fingerprint. Error detection capability is the ability of fingerprinting decoder block to report corrupted fingerprint. Bit accuracy is the proportion of correctly decoded bits to the total bits. Fingerprint error rate is the proportion of corrupted fingerprint to the total number of fingerprints.\\ *fingerprint corruption are not reported, hence, calculated in test bench with reference fingerprint.}
         \label{tab:fingerprint-accuracy}
\end{table}

\subsection{Preliminaries}
Our work is primarily designed for diffusion based models but should be applicable for GAN based models as well. The primary focus is on open source Stable Diffusion (SD)  model, that consists of $Encoder$ $\mathcal{E}: \mathbb{R}^{d_x} \rightarrow \mathbb{R}^{d_z} ; z = \mathcal{E}(x)$ which maps pixel space to a latent space, U-Net based diffusion model $\epsilon_\theta$, cross attention mechanism for textual input and $Decoder$ $\mathcal{D}:\mathbb{R}^{d_z} \rightarrow \mathbb{R}^{d_x}; x = \mathcal{D}(z)$ which maps latent space to pixel space. 

% \begin{figure}[htbp]
% \includegraphics[width=0.5\textwidth]{sec/imgs/fingerprinting_overview.png}
% \caption{Overview of the fingerprinting.}
% \label{fig:neural_fingerprinting_overview}
% \end{figure}

\subsection{Overview}
Our primary objective is to fine-tune SD's decoder such that generated artifacts contain uniquely identifiable user information without significant change in image quality and perception, this is achieved using weight modulation technique \cite{yu2022responsibledisclosuregenerativemodels, karras2020analyzingimprovingimagequality}. Our secondary objective is to train a deciphering network capable of extracting ciphered fingerprint from the generated artifacts. The overview of the neural fingerprinting architecture is shown in Fig. \ref{fig:neural_fingerprinting_overview}. A pretrained SD is used as the base, only the $decoder$ $\mathcal{D}$ is fine-tuned keeping the reset of the components untouched. 

\subsection{Fingerprint encoder}
A binary fingerprint $\phi \in \{0,1\}^{d_\phi} \sim Ber(0.5)^{d_\phi}$, of length $d_\phi$, be uniquely identifiable user information that needs to be embedded in the generated artifacts. We utilize a fingerprint encoder $\mathcal{F_E} : \mathbb{R}^{d_\phi} \rightarrow \mathbb{R}^{d_\upsilon}; \upsilon = \mathcal{F_E}(\phi)$ that is responsible for generating the ciphered fingerprint $\upsilon$. The fingerprint encoder block consists of 1) The BCH encoder $\mathcal{E_{BCH}} : \mathbb{R}^{d_\phi} \rightarrow \mathbb{R}^{d_\psi}; \psi = \mathcal{E_{BCH}}(\phi)$ based on the Bose–Chaudhuri–Hocquenghem code \cite{BOSE196068}, a class of cyclic error-correcting codes capable of detecting and correcting multi-bit errors. This block is responsible for generating unique encoded fingerprint $\psi$ from the user fingerprint $\phi$.  2) The cipher network block $\mathcal{E_{\psi}}:\mathbb{R}^{d_\psi} \rightarrow \mathbb{R}^{d_\upsilon}; \upsilon = \mathcal{E_{\psi}}(\psi)$ that generates the ciphered fingerprint $\upsilon$.

The weight modulation of the decoder $\mathcal{D}$ is achieved using StyleGAN \cite{rombach2022highresolutionimagesynthesislatent} based weight modulation, for the layer $l$,  ${W_l}' = \mathcal{A}_l({\upsilon}) * W_l$, here $\mathcal{A}_l : \mathbb{R}^{d_\upsilon} \rightarrow \mathbb{R}^{d_l}$ is the $l^{th}$ affine transformation realized using the fully connected network responsible for the dimensionality matching. The weight modulated SD decoder $\mathcal{D}: \mathbb{R}^{d_z} \rightarrow \mathbb{R}^{d_W \times d_H \times d_C}; \tilde{x_\psi} = \mathcal{D}(z)$

\begin{table}
\centering
 \setlength{\tabcolsep}{12pt}
        \resizebox{\linewidth}{!}{
        \begin{tabular}{l|cccc}
    \toprule
          Methods   & SSIM($ \uparrow $) & PSNR($ \uparrow $) & LPIPS ($ \downarrow $) & FID ($ \downarrow $)\\
    \midrule
    % WOUAF          & 0.9206 & 28.2097  & 0.0731 & 7.66681 \\
    WOUAF          & 0.9206 & 28.2097  & 0.0731 & 7.6668 \\

    \ours{}  & \textbf{0.9534}  & \textbf{30.4530}  & \textbf{0.0559} & \textbf{6.2240}\\ 

    \bottomrule
\end{tabular}
}
\caption{ A study and comparison of image quality for a 32 bit fingerprint being embedded in an image. All the metrics  are calculated w.r.t output of unmodified SD.}
\label{tab:image-quality-SD}
\end{table}

\begin{table}
\centering
 \setlength{\tabcolsep}{12pt}
        \resizebox{\linewidth}{!}{
\begin{tabular}{lllll}
\toprule
        Method & SSIM($ \uparrow $) & PSNR($ \uparrow $) & LPIPS ($ \downarrow $) & FID ($ \downarrow $)\\
    \midrule
    SD                & 0.8860 & 28.4201  & \textbf{0.1047} & \textbf{09.4744} \\
    % WOUAF             & 0.8746 & 27.1743  & 0.12078 & 13.6315 \\
    WOUAF             & 0.8746 & 27.1743  & 0.1208 & 13.6315 \\

\ours{}  & \textbf{0.9025}  & \textbf{28.4982}  & 0.1108 & 14.5790\\ 

\bottomrule
\end{tabular}
}
\caption{A study and comparison of image quality for a 32 bit fingerprint being embedded in an image. All the metrics  are calculated  w.r.t original dataset image.}
\label{tab:image-quality-OI}
\end{table}

\subsection{Fingerprint decoder}
The embedded fingerprint is extracted from the artifact using the fingerprint decoder $\mathcal{F_D}:\mathbb{R}^{d_W \times d_H \times d_C} \rightarrow \mathbb{R}^{d_\phi}$. The fingerprint decoder consists of a decipher network $\mathcal{D_\psi}:\mathbb{R}^{d_W \times d_H \times d_C} \rightarrow \mathbb{R}^{d_{\psi}}; \psi=\mathcal{D_\psi}(\tilde{x_\psi})$ that is responsible for extracting the embedded fingerprint from the artifacts, and the BCH decoder $\mathcal{D_{BCH}}:\mathbb{R}^{d_{\psi}} \rightarrow \mathbb{R}^\phi ; \phi = \mathcal{D_{BCH}}(\psi)$ based on Bose–Chaudhuri–Hocquenghem code  \cite{BOSE196068} capable of detecting, correcting, and reporting fingerprint corruption.

\begin{table*}
\begin{subtable}[t]{1\textwidth}
\centering
\setlength{\tabcolsep}{12pt}
\resizebox{0.98\linewidth}{!}{
\begin{tabular}{lllllllll}
\toprule
Method & Brightness & Contrast & Saturation & Sharpness & H Flip & G Noise & Crop & JPEG\\
         & [0.7,1.3] & [0.7,1.3] & [0.7,1.3] & [0.7,1.3] &  & $\sigma\in$[0,0.1] & [0,0.2] & q$\in$[60,99]\\
\midrule
WOUAF   & 0.9970 & 0.9969  & 0.9972 & 0.9928 & 0.9944 & 0.9991 & 0.9975
 & 0.9900\\
\ours{}  & \textbf{1.0000}  &\textbf{1.0000} & \textbf{1.0000} & \textbf{1.0000} & \textbf{1.0000} & \textbf{1.0000} & \textbf{1.0000} & \textbf{1.0000}\\ 
\bottomrule
\end{tabular}
}
\caption{The table shows the minimum fingerprint accuracy seen against different image post-processing for varied strength.}
\label{tab:acc-post-processing}
\end{subtable}

\begin{subtable}[t]{1\textwidth}
\centering
\setlength{\tabcolsep}{12pt}
\resizebox{0.98\linewidth}{!}{
\begin{tabular}{lllllllll}
\toprule
Method & Brightness & Contrast & Saturation & Sharpness & H Flip & G Noise & Crop & JPEG\\
         & [0.7,1.3] & [0.7,1.3] & [0.7,1.3] & [0.7,1.3] &  & $\sigma\in$[0,0.1] & [0,0.2] & q$\in$[60,99]\\
\midrule
WOUAF   & 0.0362 & 0.0330  & 0.0334 & 0.0842 & 0.0704 & 0.0230 & 0.0674 & 0.2235
 \\
\ours{}  & \textbf{0.0008}  &\textbf{0.0006} & \textbf{0.0010} & \textbf{0.0080} & \textbf{0.0006} & \textbf{0.0012} & \textbf{0.0112} & \textbf{0.0124}\\ 
\bottomrule
\end{tabular}
}
\caption{The table shows the maximum fingerprint error rate seen against  different image post-processing for varied strength.}
\label{tab:fer-post-processing}
\end{subtable}
\caption{ Accuracy and FER readings against varied image post-processing operations. Note accuracy drop is seen beyond above mentioned post-processing scale values, but image quality also gets affected, implying fingerprint corruption at the cost of reduced image quality.}
\end{table*}

\subsection{Loss function}
As part of the training, our primary objective is to maximize the fingerprint decoding capability of the deciphering network. This is achieved through the BCE loss function denoted as $L_\psi$ . The loss between original fingerprint $\psi$ and extracted fingerprint $\tilde{\psi}$ is 
\begin{equation}
% L_\psi(\psi, \tilde{\psi})
    L_\psi = - \sum_{i=1}^{d_\psi} \psi_i ln (\sigma(\tilde\psi_i)) + ( 1 - \psi_i) ln(1 - \sigma(\tilde\psi_i))
\end{equation}
Our secondary objective is to ensure negligible effect of fingerprint on the generated artifacts. To ensure this, we employ three loss components, the first LPIPS perceptual loss \cite{zhang2018unreasonableeffectivenessdeepfeatures} $L_{LPIPS}$ responsible for perceptual similarity  between original and generated artifacts, second SSIM loss $L_{SSIM}$ responsible for maintaining structural integrity, and third MSE loss $L_{MSE}$. Thus, secondary loss function $L_x$ is given by
\begin{equation}
    \begin{split}
%  L_x(x, \hat{x_{\psi}})
    L_x = \lambda_{1} L_{LPIPS}(x, \hat{x_{\psi}}) + \lambda_{2} L_{SSIM}(x, \hat{x_{\psi}}) + \\
    \lambda_{3} L_{MSE}(x, \hat{x_{\psi}})
    \end{split}
\end{equation}

where $\lambda_{1} = \lambda_{2} = \lambda_{3} = 1$. The total loss function is given by 
\begin{equation}
    L = L_\psi (\psi, \tilde{\psi}) + L_x(x, \hat{x_{\psi}})
\end{equation}
The ciphering network $\mathcal{E_{\psi}}$, SD decoder $\mathcal{D}$ and the deciphering network $\mathcal{D_\psi}$ are jointly optimized to ensure robust fingerprinting.

% \clearpage
% \setcounter{page}{1}
% \maketitlesupplementary

\section{Experiments}
\label{sec:results}

\subsection{Experimental Setup}
\paragraph{Dataset} We fine-tune SD decoder $\mathcal{D}$ and decipher network $\mathcal{D_{\psi}}$ to generate and decode a 512p resolution image based on MS-COCO \cite{lin2015microsoftcococommonobjects} dataset, incorporating the Karpathy split. 

\paragraph{Experimental Setup} We adopt Stable Diffusion 2.0 for fine-tuning with guidance scale 7.5 and 20 diffusion steps. Our weight modulation of SD Decoder $\mathcal{D}$ is inspired by StyleGAN2-ADA \cite{karras2020analyzingimprovingimagequality}, cipher network $\mathcal{E_{\psi}}$ is implemented using two chained fully connected network. The deciphering network $\mathcal{D_{\psi}}$ is implemented using ConvNext \cite{liu2022convnet2020s} with the last classification layer replaced by Layer Norm and FC layers, whereas WOUAF used Resnet50 and FC layer.
BCH encoder $\mathcal{E_{BCH}}$ and decoder $\mathcal{D_{BCH}}$ is configured to have a codeword length of 63 bits, message length of up to 39 bits, in this configuration it can detect and correct up to 4 bit errors in $\psi$.

\subsection{Evaluation}
\quad{} We evaluated the fine-tuned SD's decoder with the following metrics: $\mathcal{D}$ using PSNR, SSIM \cite{1284395} and LPIPS \cite{zhang2018unreasonableeffectivenessdeepfeatures}. The accuracy of fingerprint decoding is measured by 
\begin{equation}
    Acc = \frac{1}{d_{\phi}}  \sum_{i = 0}^{d_\phi} \mathcal{1}(\phi_i = \hat{\phi_i})
\end{equation}

We 
% even 
introduce and incorporate a fingerprint error rate (\textbf{FER}) metric that evaluates the number of corrupted fingerprints post decoding, formulated as 
\begin{equation}
    \textbf{FER} = \frac{Number\ of\ corrupted\ fingerprint}{Total\ number\ of\ fingerprint}  
\end{equation}

\paragraph{Fingerprint Accuracy} 
As demonstrated in Table \ref{tab:fingerprint-accuracy}, \ours{} outperforms WOUAF (our strongest baseline) significantly  and produces robust and superior performance as per all metrics. Even pre-error correction, \ours{} $\psi$'s bit accuracy is significantly higher than WOUAF. A key feature of our methodology is the ability to detect and report fingerprint corruption. 

\paragraph{Image Quality} As shown in Table \ref{tab:image-quality-SD}, the quality of generated images with our methodology is closer to the original SD's compared to WOUAF as measured by SSIM, PSNR, LPIPS, and FID. 
Table \ref{tab:image-quality-OI} shows image quality assessment for our methodology compared to Stable Diffusion and WOUAF. \ours{} performs better than WOUAF in terms of SSIM and PSNR, while improves upon SD in terms of LPIPS and FID.
Figure~\ref{fig:qualitative-results} shows qualitative results that 
highlight \ours{}'s ability to reliably incorporate fingerprinting without degrading image generation quality.

\begin{table*}[h]
\begin{subtable}[t]{1\textwidth}
\centering
 \setlength{\tabcolsep}{12pt}
    \begin{tabular}{l|ccc|ccc}
    \toprule

              & \multicolumn{3}{c|}{DDIM} & \multicolumn{3}{c}{Euler} \\ 
    \midrule
             & T = 40 & T = 50 & T = 60 & T = 15 & T = 20 & T = 25\\
    \midrule
     WOUAF    &  0.99243 & 0.99644 & 0.99036 & 0.99382 & 0.99420 &  0.99007\\
    \ours{}   &  1.00000 & 1.00000 & 1.00000 & 1.00000 & 1.00000 &  1.00000\\
    \bottomrule
    \end{tabular}
\caption{Table showing effect of different scheduler on fingerprint accuracy.}
\label{tab:sch-acc}
\end{subtable}

\begin{subtable}[t]{1\textwidth}
\centering
 \setlength{\tabcolsep}{12pt}
        \begin{tabular}{l|ccc|ccc}
    \toprule

              & \multicolumn{3}{c|}{DDIM} & \multicolumn{3}{c}{Euler} \\ 
    \midrule
             & T = 40 & T = 50 & T = 60 & T = 15 & T = 20 & T = 25\\
    \midrule
     WOUAF    &  0.17137 & 0.17741 & 0.20967 & 0.16330 & 0.07002 &  0.15725\\
    \ours{}   &  0.00980 & 0.00820 & 0.01120 & 0.01020 & 0.00720 &  0.01120\\
    \bottomrule
\end{tabular}
\caption{Table showing effect of different scheduler on fingerprint accuracy.}
\label{tab:sch-fer}
\end{subtable}
\caption{ A study of scheduler hyperparameter effect on fingerprinting }
\label{tab:sch-acc-fer}
\end{table*}

\begin{table*}[h]
\centering
 \setlength{\tabcolsep}{12pt}
        \begin{tabular}{l|ccc|ccc}
    \toprule
              & \multicolumn{3}{c|}{Accuracy} & \multicolumn{3}{c}{FER} \\
    \midrule
    guidance scale  & 5 & 7.5 & 10 & 5 & 7.5 & 10\\
    \midrule
     WOUAF    &  0.99031 & 0.99420 & 0.99784 & 0.21517 & 0.14531 & 0.16152 \\
    \ours{}   &  1.00000 & 1.00000 & 1.00000 &  0.0092 & 0.00600 & 0.00800 \\
    \bottomrule
\end{tabular}

\caption{ A study of guidance scale hyper parameter effect on fingerprint accuracy and FER. }
\label{tab:gs-acc-fer}
\end{table*}

 \begin{figure}
  \centering
  \subfloat[Plot showing bit accuracy versus PSNR for autoencoder proposed by Lee et al. \cite{lee2019contextadaptiveentropymodelendtoend}]{\includesvg[inkscapelatex=false, width = 162pt]{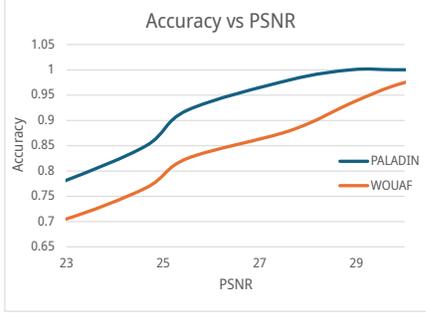}}
\vfill
  \subfloat[Plot showing bit accuracy versus PSNR for autoencoder proposed by Minnen et al. \cite{minnen2018jointautoregressivehierarchicalpriors}]
  {\includesvg[inkscapelatex=false, width = 162pt]{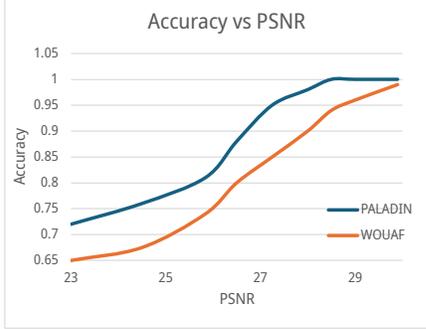}} \\
  
\caption{Plots showing bit accuracy verus PNSR when autoencoders are used to manipulate the fingerprint.}
\label{fig:acc-psnr}
\end{figure}

\subsection{Resilience against image post-processing}
\quad{} One of the most common techniques to corrupt or erase fingerprint is image post-processing techniques, in that a malicious user attempts to obscure the fingerprint by employing one or more post processing methods on the generated image whilst preserving the photo-realistic nature of image. Here, we test our methodology against multiple post processing attacks such as brightness, contrast, saturation, sharpness, horizontal flip, crop and Gaussian noise and benchmark against WOUAF. Table \ref{tab:acc-post-processing} and \ref{tab:fer-post-processing} show the minimum accuracy and maximum FER seen for a range of scale values for individual image post-processing operations, \ours{} shows better accuracy and FER against WOUAF on all image post-processing employed. It is observed that accuracy drops below 100\% beyond these range of post processing scale value, but key point to note is image quality is also affected which implies fingerprint corruption is possible at the cost of image quality. We also evaluate robustness against JPEG compression, from Table \ref{tab:acc-post-processing} and \ref{tab:fer-post-processing}  we infer that \ours{} performs better than WOUAF when employed with JPEG compression with jpeg quality ranged from 60 to 99.

\subsection{Resilience against auto encoders.}

\quad{} Deep learning techniques such as auto-encoders \cite{lee2019contextadaptiveentropymodelendtoend} \cite{minnen2018jointautoregressivehierarchicalpriors} are used with the aim of corrupting fingerprints in the artifacts. Figure \ref{fig:acc-psnr} shows the plot of accuracy versus PSNR for autoencoders \cite{lee2019contextadaptiveentropymodelendtoend} \cite{minnen2018jointautoregressivehierarchicalpriors}. We can observe \ours{} performing significantly better than WOUAF. A key point to note is that the accuracy is dropping beyond a certain PSNR value, indicating a degradation in image quality due to significant compression strength. This also suggest that the degradation in fingerprint accuracy is possible only by reducing image quality.

\section{Resilience against model hyperparameters}
\quad{}

\begin{table}[h]
\centering
 \setlength{\tabcolsep}{12pt}
        \begin{tabular}{l|c|c}
    \toprule
              & Accuracy & FER \\
    \midrule
     WOUAF    &  0.99420 & 0.01120  \\
    \ours{}   &  1.00000 & 0.00320  \\
    \bottomrule
\end{tabular}

\caption{ A study of negative prompt effect on fingerprint accuracy and FER. }
\label{tab:neg-promt-acc-fer}
\end{table}

 \begin{figure}[h]
  \centering
  \subfloat[Plot showing bit accurary for different output image sizes]{\includesvg[inkscapelatex=false, width = 180pt]{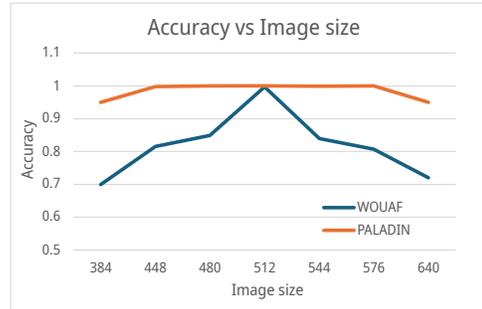}}
\hfill
  \subfloat[Plot showing FER for different output image sizes]{\includesvg[inkscapelatex=false, width = 180pt]{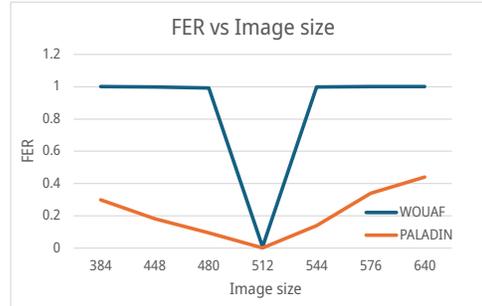}} \\
  
\caption{Plots showing accuracy and FER for varying image sizes.}
\label{fig:img_size-acc-fer}
\end{figure}

\quad In this section, model's hyperparameters such as guidance scale, number of inference steps, scheduler, and negative prompt are varied and tested to observe its effect on fingerprint accuracy and FER. \\
\quad{} Table \ref{tab:sch-acc-fer} shows the effect of using different scheduler and number of inference step on accuracy and FER. The weights used in these experiments were trained on Euler scheduler and the number of inference steps was set to 7.5. Table \ref{tab:sch-acc} shows \ours{}'s accuracy being more stable and better than WOUAF, Table \ref{tab:sch-fer} shows that our methodology has the least FER with a minimal increase in FER when inference step is varied.

\quad{} As second part, guidance scale is varied, the model was trained with the guidance scale 7.5. Table \ref{tab:gs-acc-fer} shows the affect on fingerprint accuracy and FER when the guidance scale is varied from 5 to 10, we see that both WOUAF and our methodology have little to no variation across change in guidance scale. This behaviour is partly due to guidance having its impact on denoising UNET and not on Decoder block. 

\quad{} As third part, negative prompts such as "blurry image", "low quality", etc. were used to manipulate the output of the Stable Diffusion model. Table \ref{tab:neg-promt-acc-fer} our methodology and WOUAF's accuracy do not have a significant impact in terms of accuracy and FER.

\quad{} As last part, different output image sizes are experimented, Figure \ref{fig:img_size-acc-fer} shows the impact of fingerprint accuracy and FER on different output image size. It is observed that the accuracy is maintained well in \ours{} compared to WOUAF due to the better decoding network and also because of the error correction. FER is impacted significantly for WOUAF when non-trained image sizes are selected.
\section{Conclusions}
\label{sec:conclusion}
Fingerprinting strategies for latent diffusion models, even with 99\% attribution accuracies, cannot be deployed due to the unacceptably large \emph{absolute} number of incorrect predictions involved. 
We introduce a framework, \ours{}, which preserves generated image quality and also achieves 100\% user-attribution accuracy.  
This methodology shows the performance of WOUAF can be further improved by a combination of modified fingerprint decoder architecture and a loss function better equipped to handle the trade-off between generated image quality and the attribution accuracy. 
Next, we further enhance the model performance by incorporating the ability of bit error recovery, a concept borrowed from the literature of coding theory. 
Although our methodology is built atop WOUAF, it is not restricted to this one model and provides a plug-and-play layer of cyclic error correction scheme on top of  similar models and is capable of improving user-attribution accuracy, along with an ability to appropriately flag non-identifiable fingerprints.
In this work, we also introduce a new metric FER for measuring fingerprinting accuracy, that is more accurate and informative. To the best of our knowledge, \ours{} is the first ever strategy to achieve cent percent user attribution accuracy in text-to-image diffusion models, which makes it possible for real-world deployment. 
% \quad{} \quad{}
% Figure \ref{fig:all-post-processing} shows accuracy variation versus the strength of the post processing is varied, \ours{} shows stable performance and maintains constant bit accuracy and FER over a range of post processing operation normally used to erase or corrupt fingerprint.

{
    \small
    \bibliographystyle{ieeenat_fullname}
    \bibliography{main.bib}
}


\end{document}